\documentclass[11pt]{article}

\usepackage[preprint]{acl}

\usepackage{times}
\usepackage{latexsym}

\usepackage[T1]{fontenc}

\usepackage[utf8]{inputenc}

\usepackage{microtype}

\usepackage{inconsolata}

\usepackage{graphicx}
\usepackage[most]{tcolorbox}
\usepackage{cleveref}
\usepackage{booktabs}
\usepackage{multirow}
\usepackage{bm}
\usepackage{subfigure}
\usepackage{enumitem}

\title{Mind the Gap in Cultural Alignment: \\Task-Aware Culture Management for Large Language Models}

\author{
 \textbf{Binchi Zhang\textsuperscript{1}}\thanks{Work done during an internship at NEC Laboratories America.},
 \textbf{Xujiang Zhao\textsuperscript{2}},
 \textbf{Jundong Li\textsuperscript{1}},
 \textbf{Haifeng Chen\textsuperscript{2}},
 \textbf{Zhengzhang Chen\textsuperscript{2}},
\\
 \textsuperscript{1}University of Virginia,
 \textsuperscript{2}NEC Laboratories America
}

\begin{document}
\maketitle
\begin{abstract}
Large language models (LLMs) are increasingly deployed in culturally sensitive real-world tasks. 
However, existing cultural alignment approaches fail to align LLMs' broad cultural values with the specific goals of downstream tasks and suffer from cross-culture interference.
We propose CultureManager, a novel pipeline for task-specific cultural alignment. CultureManager synthesizes task-aware cultural data in line with target task formats, grounded in culturally relevant web search results. 
To prevent conflicts between cultural norms, it manages multi-culture knowledge learned in separate adapters with a culture router that selects the appropriate one to apply.
Experiments across ten national cultures and culture-sensitive tasks show consistent improvements over prompt-based and fine-tuning baselines. 
Our results demonstrate the necessity of task adaptation and modular culture management for effective cultural alignment.
\end{abstract}

\section{Introduction}

Large language models (LLMs) have achieved remarkable success across a wide range of natural language processing tasks in various cultural contexts~\citep{rystrom2025multilingual,tao2024cultural,keleg2025llm}.  
As LLMs are deployed in applications for users with diverse cultural backgrounds, cultural alignment has emerged as an important research direction~\citep{feng2025culfit,yuan2024cultural}.
However, mainstream LLMs often reflect a Western-centric cultural bias, as their training corpora are dominated by English and other major languages~\citep{pawar2025survey,li2024culturepark}. 
This bias can lead to misinterpretation or even harm in culture-sensitive tasks such as content moderation.
For example, the ``OK'' hand sign means ``fine'' in the United States but is considered rude and offensive in Turkey.

To improve cultural awareness, previous work has explored culture-specific prompting (role play with a specific nationality)~\citep{alkhamissi2024investigating} and in-context learning with a few culturally relevant demonstrations~\citep{kim2024salad,choenni2024self}.
More broadly, researchers have fine-tuned LLMs on datasets reflecting cultural norms and values~\citep{li2024culturellm} such as the World Values Survey (WVS)~\citep{haerpfer2022world}.
Despite these advances, two critical gaps limit the implementation of fine-tuning-based approaches.

First, there is a \emph{generalization gap between cultural alignment and culture-related downstream tasks}.
Most cultural alignment datasets cover broad norms and beliefs, but are not tailored to the specific tasks where cultural sensitivity is critical. 
For example, value-oriented data (\textit{e.g.}, How important is family in your life?~\footnote{This is a real example from WVS.}) does not transfer well to offensive language detection, where cultural sensitivity can be subtle and contextual.
Given the limited availability of culturally relevant datasets, aligning cultural knowledge to task-specific formats is necessary to achieve reliable performance~\citep{zhang2020multi,chen2020low}.
Second, \emph{the multicultural interference problem} leads to a trade-off in performance across different cultures. 
In different cultural contexts, the answers to the same question can vary significantly.
Although existing work adds a culture-specific prefix to distinguish conflicting answers, multicultural alignment has shown performance degradation compared to single-culture alignment~\citep{li2024culturellm}.

\begin{figure*}[t]
    \centering
    \includegraphics[clip, trim=2cm 5.8cm 2cm 5.8cm, width=\linewidth]{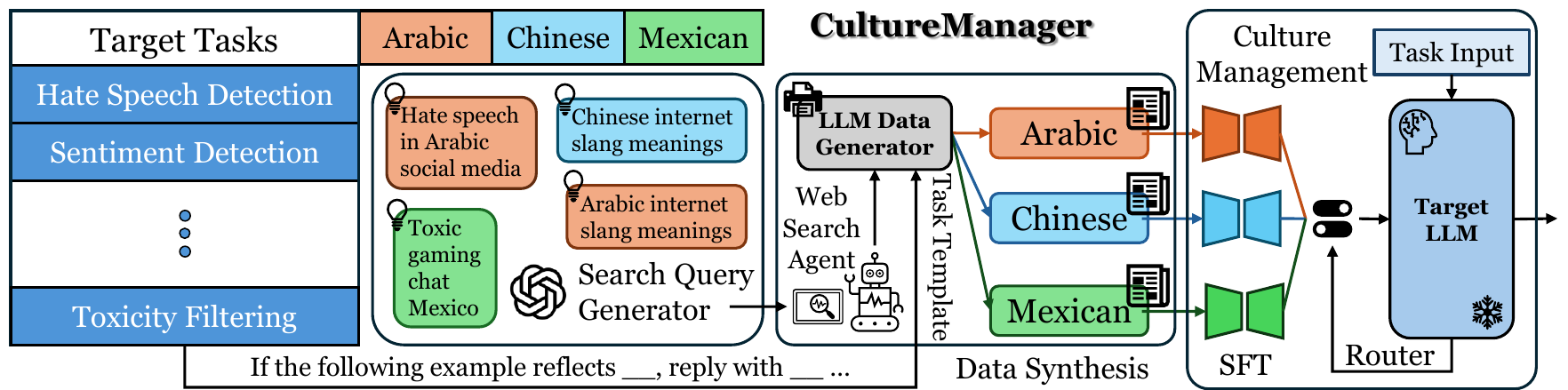}
    \vspace{-3mm}
    \caption{Overview of the CultureManager pipeline. CultureManager consists of three parts: Search Query Generation (left), Task-aware Cultural Data Synthesis (center), and Culture Management (right).}
    \label{fig:overview}
    \vspace{-3mm}
\end{figure*}

To address these limitations, we propose CultureManager, a novel task-aware cultural alignment pipeline. 
To bridge the mismatch between general cultural datasets and downstream tasks, CultureManager is given a small set of unlabeled task demonstrations and target cultures, and performs task-aware cultural data synthesis: 
an LLM-based query generator produces task-aware search queries, a web search agent retrieves culture-specific materials, and a data generator converts the retrieved content into labeled training samples in the target task format. 
The resulting synthetic data are grouped by cultures to build culture-specific training sets. 
For each culture, CultureManager fine-tunes a lightweight LoRA adapter~\citep{hu2022lora} while keeping the base LLM frozen. 
During inference, a culture router selects the appropriate adapter based on the input cultural context to produce culturally aligned outputs. 
We evaluate CultureManager on five national cultures and ten culture-sensitive tasks, showing consistent improvements and validating the effectiveness of task-aware data synthesis and modular culture management.
Our study demonstrates that under limited data sources, culture-sensitive applications benefit from task adaptation and culture management.
Our contributions are threefold:
\begin{itemize}[leftmargin=*]
\item We identify two fundamental limitations of existing cultural alignment methods:
(i) a generalization gap between broad cultural alignment data and culture-sensitive downstream tasks, and
(ii) cross-culture interference that degrades performance in multicultural settings.

\item We propose CultureManager, a task-aware and modular cultural alignment pipeline that synthesizes task-specific cultural training data from culturally grounded web sources, and isolates multicultural knowledge into lightweight, culture-specific adapters with dynamic routing.

\item We conduct extensive experiments across five national cultures and ten culture-sensitive tasks, demonstrating consistent improvements over prompt-based and fine-tuning baselines, and showing that task adaptation and modular culture management are crucial for effective cultural alignment.
\end{itemize}

\section{Related Work}
Recent work has shown that LLMs inherit and amplify cultural patterns present in their training data~\citep{li2025understanding,sukiennik2025evaluation,masoud2025cultural,qiu2025evaluating}.
Models have been shown to replicate cross-cultural personality differences~\citep{niszczota2025large} and overrepresent Western perspectives due to English-dominant corpora~\citep{kim2025dual,li2024well,mushtaq2025worldview}.
These findings motivate efforts to increase cultural awareness and mitigate cultural bias in LLM outputs.
An efficient strategy is to align outputs with a specific culture through prompting.
Early work conceptualizes LLMs as superpositions of cultural viewpoints that can be activated through prompting~\citep{kovavc2023large,zhong2024cultural}.
Persona prompting~\citep{alkhamissi2024investigating,masoud2024llm} and in-context learning~\citep{kim2024salad,choenni2024self} have also been explored.
While prompt-based methods are efficient, they depend on cultural representations already present in the model and degrade for low-resource cultures with limited representations~\citep{alhanai2025bridging}.
Another line of work directly embeds cultural knowledge into model parameters through fine-tuning~\citep{feng2024modular,yuan2024cultural}.
Researchers have leveraged structured surveys such as WVS~\citep{haerpfer2022world,mushtaq2025worldview} and curated web or social content~\citep{chiu2025culturalbench,myung2024blend,shi2024culturebank}.
These methods improve cultural alignment but remain challenging to scale, as collecting high-quality cultural data is difficult, especially for low-resource cultures.
To mitigate data scarcity, recent work explores synthesizing culture-specific data using LLMs~\citep{xu2025self,li2024culturepark}.
However, most synthesis efforts mimic the distribution of general cultural knowledge, leaving a task adaptation gap when applying models in practice.

\section{Methodology}
The overall pipeline of CultureManager is illustrated in~\Cref{fig:overview}. 
Next, we introduce each module of CultureManager in detail.

\paragraph{Preliminary}
The first step of CultureManager is to determine the target downstream tasks and cultures.
In this paper, we see culture as a set of shared norms and values, and cultural alignment as modeling the conditional distribution of language under a specific cultural context~\citep{fung2023normsage,rao2025normad,sukiennik2025evaluation,alkhamissi2024investigating,yao2025caredio}, emphasizing culturally grounded knowledge and practices. 
The input of CultureManager includes a set of task-level labels $\mathcal{T}$ (\textit{e.g.}, ``hate speech detection''), a small set of unlabeled task demonstrations $\mathcal{D}_t$ for each task $t\in\mathcal{T}$, and a set of cultures $\mathcal{C}$.
We denote that $\mathcal{D}=\{\mathcal{D}_t\}_{t\in\mathcal{T}}$.
The format of task demonstrations should align with the task inputs during inference.
This is a practical setting when LLMs are implemented to perform specialized tasks~\citep{ling2023domain,lin2025investigating}.

\paragraph{Search Query Generation}
We set up two modes of search query generation: task-specific and task-agnostic modes, to capture task-specific cultural knowledge and broader cultural knowledge.
In particular, we prompt the GPT-4o model~\citep{hurst2024gpt} to generate search queries.
The prompt templates are provided in~\Cref{sec:prompt}.
The inputs of the query generator are: number of generated queries $n$, culture label $c$, task label $t$, and task demonstration batch $\mathcal{B}_t\subseteq\mathcal{D}_t$.
We iterate over all cultures $c\in\mathcal{C}$ and tasks $t\in\mathcal{T}$ and randomly sample $\mathcal{B}_t\sim\mathcal{D}_t^b$ ($b$ denotes the batch size) each time, resulting in $n|\mathcal{C}|\cdot|\mathcal{T}|$ queries in total.

\paragraph{Data Synthesis}
We leverage a lightweight implementation of SearchGPT~\footnote{\url{https://github.com/Wilson-ZheLin/SearchGPT}} as the web search agent module.
For each generated query, the search agent retrieves the top-$k$ webpage through the Google search engine and summarizes the returned content with GPT-4o.
The faithfulness and trustworthiness of the retrieved content are guaranteed by the alignment of the adopted GPT model~\citep{li2024culturepark,yao2025caredio}.
For data synthesis, we still apply GPT-4o as the data generator. 
The prompt template for data synthesis is provided in~\Cref{sec:prompt}.
The inputs of the data generator are: number of generated samples $m$, culture label $c$, task label $t$, task demonstration batch $\mathcal{B}_t$, and the retrieved material.
We denote the generated data samples for culture $c$ and task $t$ as $\mathcal{S}_{c,t}$ and have $|\mathcal{S}_{c,t}|=m$.
After data synthesis, we collect the samples under each culture as $\mathcal{S}_c=\cup_{t\in\mathcal{T}}\mathcal{S}_{c,t}$ for further cultural alignment.
The total number of generated data is $nm|\mathcal{C}|\cdot|\mathcal{T}|$.

\paragraph{Cultural Alignment and Management}
The synthetic datasets $\mathcal{S}_c$ generated in the last step are well-formatted and can be used for fine-tuning.
We denote $f_\Theta$ as the target LLM parameterized by $\Theta$ and denote $\Theta_c$ as the parameters of the LoRA adapter corresponding to culture $c$.
We fine-tune the LoRA adapter $\Theta_c$ as follows
\begin{equation}
\min_{\Theta_c}\mathrm{E}_{(x,y)\in\mathcal{S}_c}\mathcal{L}\left(f_{\Theta\cup\Theta_c}(x), y\right),
\end{equation}
where $\mathcal{L}()$ is the cross-entropy loss for standard question-answering tasks.
After fine-tuning for all cultures, we obtain a set of culturally aligned LoRA adapters $\{\Theta_c\}_{c\in\mathcal{C}}$.

During inference, a task input $x_t$ is fed into the culture router, which is instantiated by prompting the target LLM $f_\Theta$:
``You are a helpful chatbot that knows different cultures around the world very well. Your task is to analyze the provided text based on its language, expressions, and context, and select the most relevant culture to the provided text from the following options:
- \{\texttt{Culture1}\}
- \{\texttt{Culture2}\}
- ...
- Others (if the text is not relevant to any cultures listed above)
Output **only** the exact name of the culture without any explanation. 
Text: \{\texttt{input}\}
Answer:''
A corresponding LoRA adapter is selected based on the router's response and incorporated with the target LLM to generate the task output.
If the answer is ``Others'', none of the LoRA adapters are activated.
Compared to learning-based routers~\citep{wang2024wise,zhang2025resolving}, our method is more robust.

\begin{table*}[!t]
\small
\renewcommand{\arraystretch}{1.05}
\tabcolsep = 15pt
\centering
\caption{Experimental results of F1 scores on different culture-sensitive downstream tasks. ``ar'', ``bn'', ``zh'', ``es'', ``tr'' are abbreviations of ``Arabic'', ``Bengali'', ``Chinese'', ``Spanish'', and ``Turkish''. The ``average'' column is computed by averaging the task metrics under the corresponding culture. For training-based methods, we record the mean and standard deviation of metrics across five runs with different random seeds. We bold the highest score and underline the second-highest score.}
\label{tab:main_results}
\vspace{-3mm}
\aboverulesep = 0pt
\belowrulesep = 0pt
\begin{tabular}{l|ccc|c}
\toprule
Method & \multicolumn{3}{c|}{CultureLLM Benchmark} & CulturalBench \\
\midrule
& ar-hate & ar-offensive & ar-average & China \\
\midrule
Original & 23.21 & 38.72 & 30.97 & 36.07 \\
Prompt & 36.53 & \textbf{76.15} & 56.34 & 25.00 \\
TaskSFT & 26.60 $\pm$ 2.45 & 63.50 $\pm$ 0.86 & 45.05 & 45.41 $\pm$ 1.16 \\
CultureSFT & 33.11 $\pm$ 5.98 & 52.79 $\pm$ 1.60 & 42.95 & 42.54 $\pm$ 1.15 \\
CultureSFT-all & 32.77 $\pm$ 4.31 & 42.41 $\pm$ 4.09 & 37.59 & 44.04 $\pm$ 1.22 \\
CultureLLM & 35.86 $\pm$ 3.46 & 63.60 $\pm$ 1.60 & 49.73 & 47.63 $\pm$ 1.99 \\
CultureLLM-all & \underline{44.45 $\pm$ 14.04} & 68.84 $\pm$ 7.42 & \underline{56.64} & \underline{53.30 $\pm$ 1.89} \\
CultureManager & \textbf{44.98 $\bm\pm$ 1.27} & \underline{69.39 $\pm$ 0.84} & \textbf{57.18} & \textbf{53.40 $\bm\pm$ 6.83} \\
\midrule
& bn-racism & bn-threat & bn-average & Germany \\
\midrule
Original & 50.69 & 45.66 & 48.18 & 30.30 \\
Prompt & 49.35 & \underline{49.31} & 49.33 & 20.00 \\
TaskSFT & \underline{52.13 $\pm$ 1.22} & 44.43 $\pm$ 6.75 & 48.28 & 40.00 $\pm$ 0.00 \\
CultureSFT & 51.93 $\pm$ 0.62 & 46.39 $\pm$ 1.33 & 49.16 & 45.50 $\pm$ 12.34 \\
CultureSFT-all & \textbf{53.12 $\bm\pm$ 2.36} & 43.52 $\pm$ 3.35 & 48.32 & 45.10 $\pm$ 8.91 \\
CultureLLM & 51.48 $\pm$ 0.92 & 41.11 $\pm$ 1.99 & 46.30 & \underline{59.20 $\pm$ 3.20} \\
CultureLLM-all & 51.01 $\pm$ 1.26 & 47.65 $\pm$ 2.03 & \underline{49.33} & 52.44 $\pm$ 6.64 \\
CultureManager & 51.37 $\pm$ 1.62 & \textbf{51.70 $\bm\pm$ 2.00} & \textbf{51.53} & \textbf{62.91 $\bm\pm$ 3.56} \\
\midrule
& zh-bias & zh-spam & zh-average & Korea \\ 
\midrule
Original & 16.67 & 43.25 & 29.96 & 38.30 \\
Prompt & 0.00 & 55.25 & 27.63 & 40.00 \\
TaskSFT & 20.23 $\pm$ 6.75 & 41.29 $\pm$ 0.25 & 30.76 & 48.47 $\pm$ 1.39 \\
CultureSFT & 8.22 $\pm$ 0.70 & 37.85 $\pm$ 0.24 & 23.04 & \underline{49.24 $\pm$ 5.21} \\
CultureSFT-all & 8.52 $\pm$ 3.91 & 38.72 $\pm$ 1.89 & 23.62 & 47.14 $\pm$ 4.38 \\
CultureLLM & 16.97 $\pm$ 12.03 & \underline{46.60 $\pm$ 0.79} & 31.78 & 47.71 $\pm$ 3.96 \\
CultureLLM-all & \underline{35.54 $\pm$ 14.56} & 37.33 $\pm$ 2.47 & \underline{36.44} & 46.82 $\pm$ 2.09 \\
CultureManager & \textbf{35.63 $\bm\pm$ 14.89} & \textbf{63.51 $\bm\pm$ 3.98} & \textbf{49.57} & \textbf{60.71 $\bm\pm$ 0.81} \\
\midrule
& es-stance & es-stereotype & es-average & Spain \\ 
\midrule
Original & 50.68 & 46.09 & 48.39 & 58.33 \\
Prompt & 49.68 & 55.52 & 52.60 & 45.00 \\
TaskSFT & \underline{54.36 $\pm$ 2.63} & \underline{55.91 $\pm$ 2.90} & \textbf{55.14} & 55.72 $\pm$ 0.69 \\
CultureSFT & 44.46 $\pm$ 1.15 & 55.11 $\pm$ 0.87 & 49.79 & 48.78 $\pm$ 0.00 \\
CultureSFT-all & 42.74 $\pm$ 3.71 & 48.13 $\pm$ 1.67 & 45.44 & 64.38 $\pm$ 3.55 \\
CultureLLM & 49.16 $\pm$ 2.82 & \textbf{57.62 $\bm\pm$ 0.70} & 53.39 & 65.41 $\pm$ 1.41 \\
CultureLLM-all & 48.06 $\pm$ 2.39 & 53.18 $\pm$ 1.95 & 50.62 & \textbf{68.15 $\bm\pm$ 2.68} \\
CultureManager & \textbf{58.08 $\bm\pm$ 2.47} & 51.35 $\pm$ 1.51 & \underline{54.71} & \underline{66.66 $\pm$ 0.99} \\
\midrule
& tr-abusive & tr-spam & tr-average & Turkey \\ 
\midrule
Original & 50.47 & 37.80 & 44.14 & 47.83 \\
Prompt & \underline{69.22} & 45.62 & 57.42 & 27.78 \\
TaskSFT & 50.70 $\pm$ 6.55 & 42.96 $\pm$ 1.10 & 46.83 & 45.90 $\pm$ 0.00 \\
CultureSFT & 62.97 $\pm$ 1.88 & 42.24 $\pm$ 0.72 & 52.61 & \underline{50.08 $\pm$ 4.04} \\
CultureSFT-all & 48.28 $\pm$ 3.19 & 40.67 $\pm$ 2.34 & 44.48 & 48.26 $\pm$ 0.87 \\
CultureLLM & 65.80 $\pm$ 0.54 & \underline{51.24 $\pm$ 2.57} & \underline{58.52} & 45.77 $\pm$ 1.15 \\
CultureLLM-all & 68.55 $\pm$ 10.87 & 47.21 $\pm$ 1.27 & 57.88 & 49.34 $\pm$ 1.88 \\
CultureManager & \textbf{70.40 $\bm\pm$ 3.78} & \textbf{52.97 $\bm\pm$ 1.36} & \textbf{61.69} & \textbf{56.55 $\bm\pm$ 1.79} \\
\bottomrule
\end{tabular}
\end{table*}

\begin{table*}[t]
\centering
\caption{Statistics of downstream task datasets.}
\label{tab:datasets}
\vspace{-3mm}
\aboverulesep = 0pt
\belowrulesep = 0pt
\begin{tabular}{ccc|cc}
\toprule
\multicolumn{3}{c|}{CultureLLM Benchmark} & \multicolumn{2}{c}{CulturalBench} \\
\midrule
Culture & Task & Labels (1 / 0) & Culture & Labels (1 / 0) \\
\midrule
\multirow{2}{*}{Arabic} & hate speech detection & 180 / 495 & \multirow{2}{*}{Chinese} & \multirow{2}{*}{62 / 174} \\
& offensive language detection & 402 / 1,598 \\
\multirow{2}{*}{Bengali} & racism detection & 48 / 951 & \multirow{2}{*}{German} & \multirow{2}{*}{35 / 93} \\
& threat detection & 70 / 929 \\
\multirow{2}{*}{Chinese} & gender bias detection & 72 / 928 & \multirow{2}{*}{Korean} & \multirow{2}{*}{44 / 120} \\
& spam detection & 482 / 518 \\
\multirow{2}{*}{Spanish} & negative stance detection & 148 / 852 & \multirow{2}{*}{Spanish} & \multirow{2}{*}{41 / 119} \\
& stereotype detection & 109 / 891 \\
\multirow{2}{*}{Turkish} & abusive language detection & 349 / 651 & \multirow{2}{*}{Turkish} & \multirow{2}{*}{38 / 114} \\
& spam detection & 332 / 493 \\
\bottomrule
\end{tabular}
\end{table*}

\section{Experiments}
In this section, we evaluate the performance of CultureManager in multicultural downstream tasks.
Through experimental results, we aim to answer these research questions:
RQ1: Can task-aware cultural alignment improve task performance better than broad cultural value alignment?
RQ2: Can CultureManager improve the model performance on culture-sensitive downstream tasks?
RQ3: Can task-specific data synthesis bridge the gap between cultural alignment and task adaptation?
RQ4: Can culture management mitigate cross-cultural interference when tackling multiple cultures?

\subsection{Datasets}
We leverage two cultural benchmarks of culture-sensitive downstream tasks to evaluate the cultural alignment baselines: CultureLLM~\citep{li2024culturellm} and CulturalBench~\citep{chiu2025culturalbench}.
We provide the statistics of downstream tasks in~\Cref{tab:datasets}.

For CultureLLM benchmarks, we chose five cultures and ten culture-sensitive downstream tasks (two for each culture), which are listed below.
Arabic: offensive language detection~\citep{zampieri2020semeval} and hate speech detection~\citep{chowdhury2020multi};
Bengali: threat detection and racism detection~\citep{li2024culturellm,aimansnigdha2018bangla};
Chinese: spam detection~\citep{jiang2019detect} and gender bias detection~\citep{zhou2022towards};
Spanish: stereotype detection and negative stance detection~\citep{magnossao2021ai};
Turkish: abusive language detection~\citep{karayiugit2021detecting} and spam detection~\citep{uci2019turkish}.
For each downstream task, we provide a system prompt before presenting the data samples to LLM:
``If the following sentence has \{\texttt{task}\}, respond with `1'. If not, respond with `0'. Do not provide any explanation, reasoning, or extra words. Sentence: \{\texttt{input}\} Response: ''.

CulturalBench is designed to rigorously assess models’ cultural knowledge across a diverse set of regions and topics, which comprises 1,696 human-written and human-verified questions covering 45 global regions, and spans 17 culturally salient topics ranging from food preferences to social etiquette. 
Each question has been verified by multiple independent annotators to ensure cultural validity and robustness. 
To provide a nuanced evaluation, CulturalBench supports two evaluation formats: an Easy multiple-choice setup and a Hard true/false setup where multiple correct answers must be identified, making it significantly more challenging for LLMs. 
In our experiment, we also chose five most frequent cultures from the CulturalBench-Hard split as the downstream tasks: China, Germany, South Korea, Spain, and Turkey. 
We use the system prompt as 
``
Question: \{\texttt{Question}\}
Answer: \{\texttt{Answer}\}
Is this answer true or false for this question? You must choose either True or False.
''

We use F1 score as the evaluation metric for all downstream tasks in the two selected benchmarks.
For baseline methods (cultural value alignment), we adopted the WVS-7 (2017-2022) dataset, version 5.0~\citep {haerpfer2022world}, spanning from mid-2017 to the end of 2021, as the training dataset.
WVS has 294 questions in total, and we followed~\citep{li2024culturellm} to rewrite 50 questions into the question-answer format. 
The rewriting process is to convert the multi-choice questions from WVS into QA format. 
For example, the original question ``Do you agree with One of my main goals in life has been to make my parents proud?'' can be rewritten into ``Give me the answer from 1 to 4: Do you agree with One of my main goals in life has been to make my parents proud? 1. Strongly agree 2. Agree 3. Disagree 4. Strongly disagree. You can only choose one option.''~\citep{li2024culturellm}.
We preprocess all these data samples following~\citep{li2024culturellm}. 

\begin{figure*}[t]
    \centering
    \includegraphics[width=0.75\linewidth]{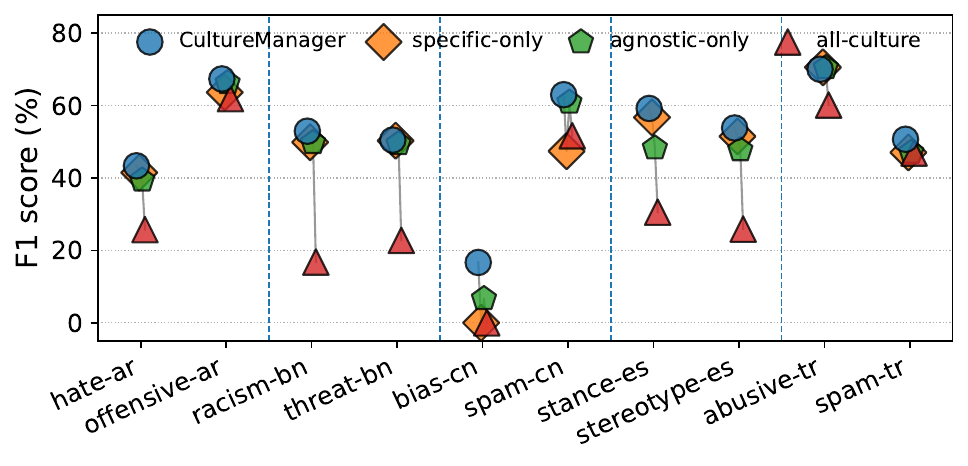}
    \vspace{-5mm}
    \caption{Experimental results of the ablation study. ``specific-only'': use only task-specific queries in data synthesis; ``agnostic-only'': use only task-agnostic queries in data synthesis; ``all-culture'': abandon knowledge management and train the model on all datasets.}
    \label{fig:ablation_study}
\end{figure*}

\subsection{Baselines}
We use Llama-3.1-8B-Instruct~\citep{dubey2024llama} as the target LLM for CultureLLM Benchmark and Qwen-2.5-7B-Instruct~\citep{bai2023qwen} as the target LLM for CulturalBench.
We select the following baselines for cultural alignment:
\begin{itemize}[leftmargin=*]
\item Original: the original target LLM; 
\item Prompt: the original model (without explicit cultural alignment) using anthropological prompting~\citep{alkhamissi2024investigating} as the system prompt. We provide the detailed anthropological prompt template in~\Cref{sec:prompt};
\item TaskSFT: the model fine-tuned directly on each downstream task. We use the same train/val split as CultureManager in the training of TaskSFT for each task dataset;
\item CultureSFT: the model fine-tuned on the WVS dataset corresponding to a specific culture;
\item CultureSFT-all: the model fine-tuned on the combination of WVS datasets for all 5 cultures;
\item CultureLLM: the model fine-tuned on the augmented dataset~\citep{li2024culturellm} from the WVS dataset corresponding to a specific culture;
\item CultureLLM-all: the model fine-tuned on the combination of augmented datasets~\citep{li2024culturellm} from all WVS datasets;
\end{itemize}
For CultureManager, we select 10\% input data of each downstream task $t$ as $\mathcal{D}_t$. For a fair comparison, we set the number of synthetic data samples to match the data augmentation method used in CultureLLM (around 1,000 per culture).
For the single-culture setting, we make an additional assumption: the downstream task is associated with a single and known culture, allowing us to fine-tune the model for each culture and evaluate the model on the corresponding tasks.
In~\Cref{tab:main_results}, we present this ideal setup to illustrate the impact of cross-cultural interference and demonstrate the effectiveness of culture management in CultureManager.

\subsection{Main Results}
As shown in~\Cref{tab:main_results}, we can observe that
(1) CultureManager achieves desirable and robust performance across all cultures (RQ2).
(2) CultureManager (task-specific cultural alignment) outperforms CultureSFT and CultureLLM (broad cultural value alignment), indicating the effectiveness of task-aware data synthesis (RQ1 and RQ3).
(3) Comparing single-culture fine-tuning and multi-culture fine-tuning, we can observe that extending the training dataset from single culture to multiple cultures does not always yield a better performance, validating the existence of cross-culture interference. CultureManager outperforms existing single- or multi-culture fine-tuning methods across most downstream tasks, indicating the effectiveness of the culture management mechanism in mitigating cross-culture interference (RQ4).
(4) CultureManager outperforms TaskSFT, which directly fine-tunes the target LLM for each downstream task, demonstrating the necessity of cultural alignment for culture-sensitive tasks. 
(5) Prompt-based methods fail to consistently improve the performance across all cultures, revealing the shortness of cultural knowledge of existing LLMs and necessitating fine-tuning for low-resource cultural alignment.

\subsection{Ablation Study}
We conducted ablation studies on the modules of CultureManager. 
We removed the task-agnostic query generation, the task-specific query generation, and the culture management modules from CultureManager, and reported the task performances on CultureLLM Benchmark in~\Cref{fig:ablation_study}.
The results demonstrate that:
(1) Combining task-specific cultural knowledge and task-agnostic cultural knowledge ensures a robust performance across all cultures (specific-only performed badly on Chinese culture and agnostic-only failed on Spanish culture).
(2) Culture management is crucial for task-aware cultural alignment, especially when different tasks are involved in different cultures. The performance drop of the all-culture setting compared with the single-culture setting is even larger than the cultural value alignment in~\Cref{tab:main_results} as diverse task templates aggravate cross-cultural interference.
(3) On most tasks (except Spanish), agnostic-only achieves comparable performance to CultureManager, providing a strong variant when task demonstrations are unavailable (still requires a task template for data synthesis).

\subsection{Data Synthesis Examples}
We present synthetic data samples for the CultureLLM Benchmark and CulturalBench in this section.
Although all queries are generated in English, if the retrieved materials and downstream tasks are in a foreign language, the generated data samples can be formatted in that language as well.
This results in naturally multilingual data synthesis, which has been shown beneficial for cultural alignment~\citep{choenni2024self}.

To mitigate data leakage risks, we strictly held out the test datasets from the synthetic data generation pipeline. 
Our web queries only reflect general cultural topics and task formats, without memorizing benchmark-specific prompts or known example instances. 
Additionally, we verified that there was no verbatim overlap between the synthetic and test examples.
We performed a hash-based deduplication check~\citep{lee2022deduplicating,wang2024mathpile} between our synthetic training data and all evaluation splits, using the sha1 hash function. 
Results showed that there are no exact duplicates in the synthetic training data, confirming that the synthetic data does not unintentionally replicate benchmark examples verbatim.

\begin{tcolorbox}[colback=white, colframe=blue!50, title=Synthetic data samples for CultureLLM Benchmark by CultureManager]
\textbf{Query (negative stance detection)}: ``Spanish cultural expressions of negative stance in social media''

\textbf{Sample}: ``Durante la Semana del Orgullo, las redes están llenas de odio hacia la comunidad LGTBIQ+. Es triste ver tanta intolerancia en un evento que solo busca celebrar la diversidad.''

\textbf{Query (gender bias detection)}: ``Impact of Chinese traditional values on gender perceptions''

\textbf{Sample}: ``In rural villages, it's common for parents to prioritize their sons' education over their daughters'. Girls often face the expectation to leave school early to help with household chores or get married. This mindset significantly limits their social mobility and perpetuates gender inequality.''
\end{tcolorbox}
\begin{tcolorbox}[colback=white, colframe=blue!50, title=Synthetic data samples for CulturalBench by CultureManager]
\textbf{Query (Germany)}: ``How do German cultural values influence workplace environments and practices?''

\textbf{Sample}: 
``\texttt{question}: In German corporate culture, how important is adherence to rules and regulations?
\texttt{answer}: Rules and regulations are strictly followed and are a cornerstone of workplace behavior.
\texttt{label}: True''

\textbf{Query (Korea)}: ``Traditional Korean family values and their impact on modern society''

\textbf{Sample}: ``
\texttt{question}: How do traditional Korean values view the role of family in education?
\texttt{answer}: Parents are expected to be actively involved in their children's education to help them succeed.
\texttt{label}: True''
\end{tcolorbox}


\begin{figure*}[t]
\centering
\subfigure[Llama-3.1-8B-Instruct on CultureLLM Benchmark]{\includegraphics[width=0.45\linewidth]{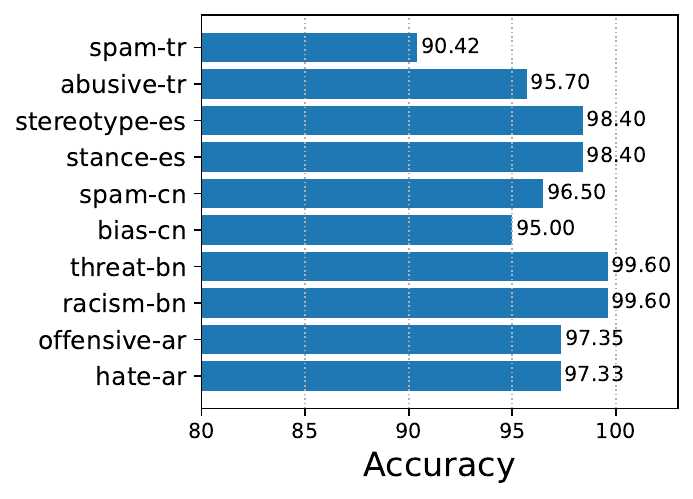}}
\subfigure[Qwen-2.5-7B-Instruct on CulturalBench]{\includegraphics[width=0.45\linewidth]{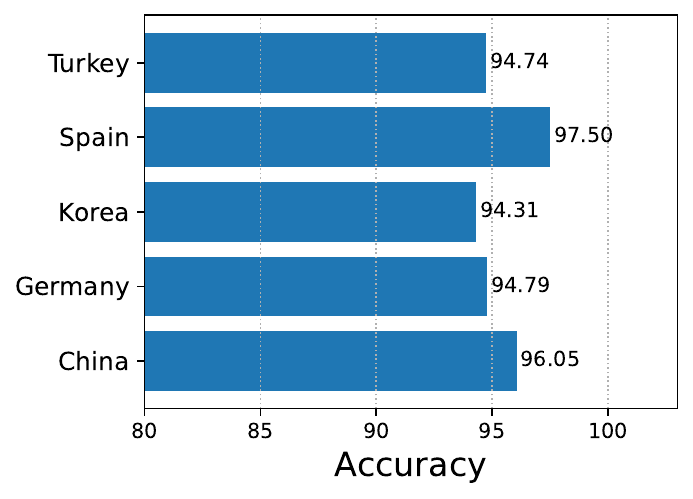}}
\vspace{-3mm}
\caption{Accuracy of the culture router on different culture-sensitive tasks.}
\label{fig:router_acc}
\end{figure*}

\subsection{Culture Router Utility}
During inference, CultureManager uses a culture router to select the most relevant culture based on the task input.
We evaluate the utility of the culture router in this experiment.
We feed each task dataset into the culture router and record its accuracy in determining the culture of the input data.
The results are shown in~\Cref{fig:router_acc}.
We observe that the culture router achieves accuracy exceeding 95\% across most tasks, indicating its effectiveness in routing the task input to the corresponding culture.
The relatively low performance in spam detection might be because spam samples are less strongly relevant to culture than those in other tasks.
Additionally, the culture router is based on Llama-3.1-8B-Instruct and Qwen-2.5-7B-Instruct. We can also leverage more powerful LLMs in practice to improve the utility of the culture router.

\begin{figure}[t]
\centering
\includegraphics[width=\linewidth]{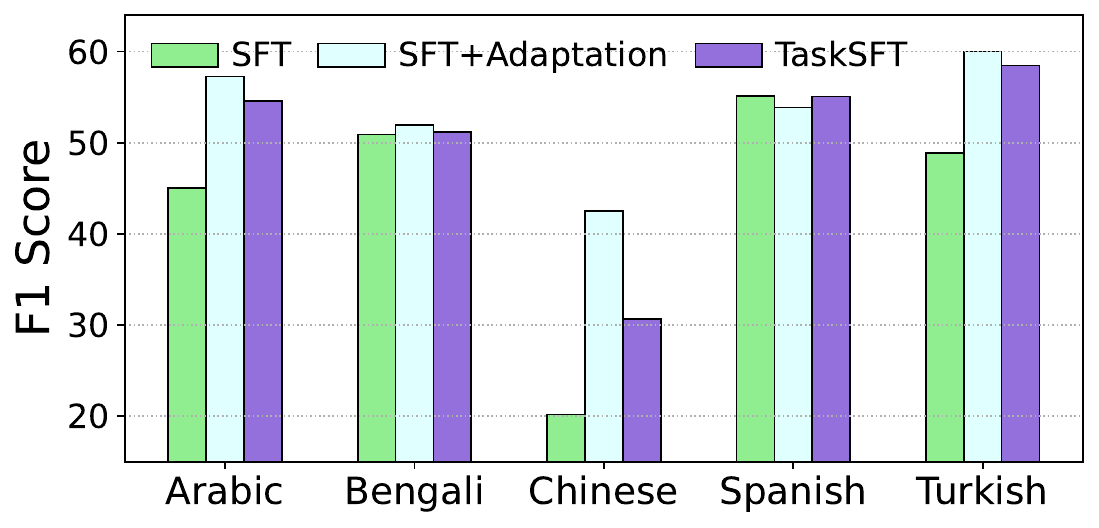}
\caption{Results of task adaptation on different cultures. SFT represents CultureSFT; SFT+Adaptation means firstly fine-tuning the LLM using CultureSFT and then continually fine-tuning it using the synthetic data; TaskSFT denotes directly fine-tuning the original LLM using the synthetic data. The metric of each culture is computed as the mean value across the tasks under that culture.}
\label{fig:task_adaptation}
\end{figure}

\subsection{Task Adaptation}
We have validated the need for task adaptation to achieve LLM cultural alignment for culture-sensitive downstream tasks.
In this section, we explore combining CultureSFT with task-aware cultural alignment in a two-stage pipeline: we first fine-tune the model on broad cultural value datasets and then continue fine-tuning on synthetic task datasets.
We compare the two-stage task adaptation pipeline with CultureSFT, denoted as SFT, and directly fine-tuning the original LLM on synthetic task datasets (as in CultureManager), denoted as TaskSFT.
The results in~\Cref{fig:task_adaptation} demonstrate that the explicit task adaptation outperforms CultureSFT and TaskSFT on four out of all five cultures.
This study sheds light on the potential for improving cultural alignment in a two-stage task-adaptation pipeline. 
In addition, improving the robustness of task adaptation and reducing conflicts between broad cultural value alignment and task-aware cultural alignment can be important research directions for future studies.

\begin{table}[t]
\small
\renewcommand{\arraystretch}{1.1}
\centering
\caption{Experimental results of culture-sensitive downstream tasks with Llama-3.3-70B-Instruct.}
\label{tab:exp_70b}
\vspace{-3mm}
\aboverulesep = 0pt
\belowrulesep = 0pt
\begin{tabular}{l|ccc}
\toprule
Method & ar-hate & ar-offensive & ar-average \\
\midrule
Original & 53.34 & 76.88 & 65.11 \\
Prompt & 52.58 & 81.75 & 67.17 \\
CultureSFT & 52.66 & 73.91 & 63.29 \\
CultureLLM & 53.38 & 72.13 & 62.76 \\
CultureManager & 55.89 & 79.33 & 67.61 \\
\midrule
& bn-racism & bn-threat & bn-average \\
\midrule
Original & 35.28 & 30.74 & 33.01 \\
Prompt & 36.38 & 0.00 & 18.19 \\
CultureSFT & 40.82 & 33.61 & 37.22 \\
CultureLLM & 37.83 & 45.08 & 41.46 \\
CultureManager & 51.16 & 50.29 & 50.73 \\
\bottomrule
\end{tabular}
\end{table}

\subsection{Model Scale}
In~\Cref{tab:main_results}, we evaluate cultural alignment on Llama-3.1-8B-Instruct~\citep{dubey2024llama}.
To explore the effectiveness of CultureManager under scale, we switch the target model to Llama-3.3-70B-Instruct~\citep{dubey2024llama}.
We select two cultures and vanilla baselines and demonstrate the results in~\Cref{tab:exp_70b}.
We can observe that
(1) CultureManager still achieves a desirable performance across both cultures.
(2) The 70B model outperforms the 8B model on Arabic, but underperforms the 8B model on Bengali, indicating that scaling is not equivalent to improved cultural awareness.

\section{Conclusion}
We introduced CultureManager, a task-aware cultural alignment pipeline for LLMs under limited cultural resources. 
CultureManager combines task-aware cultural data synthesis with a modular culture management strategy to bridge the gap between broad cultural knowledge and task-specific nuances and resolve cross-cultural interference.
Experiments across five national cultures and ten culture-sensitive downstream tasks demonstrate consistent improvements over prompting and cultural value alignment baselines, validating both the task adaptation and culture management components. 
These results indicate that cultural alignment cannot rely solely on broad cultural value data or one-for-all models, but should integrate downstream task needs and explicitly handle cultural divergence.

\section*{Limitations}

Our study primarily focuses on predefined downstream tasks (\textit{e.g.}, offensive language and hate speech detection) and a fixed set of national cultures. 
This scope enables controlled evaluation and precise measurement of improvements, but does not preclude broader use. 
The design of CultureManager can be easily adapted to task-agnostic settings: removing the task-template input in the data synthesis module would allow the same pipeline to generalize to unseen tasks. 
In addition, our experimental settings did not include tasks that jointly involve multiple cultures, such as culture-sensitive translation. 
These tasks require modeling more complex, cross-cultural semantics and routing across multiple cultural components simultaneously. 
While not the focus of this work, the modular management design of CultureManager naturally extends to multi-hop or mixed-culture inference, which we plan to explore in future work.
From the resource perspective, our approach proposed a fully automated data synthesis pipeline, without explicit human validation of individual synthetic examples. 
While human review could improve the precision and factual correctness of the generated data, it would substantially reduce the scalability and efficiency that are central to our method’s design. 
We view this trade-off as intentional: our approach aims to enable efficient, large-scale data augmentation with minimal manual intervention. 
Incorporating selective human validation or hybrid human-in-the-loop mechanisms remains an interesting direction for future work.

\section*{Ethical Considerations}

This work focuses on improving the cultural alignment of LLMs to reduce unintended harm in culture-sensitive applications. 
All cultural data used is derived from publicly available web sources without collecting private or toxic information.
The method is designed to support safer, more inclusive NLP systems without introducing additional ethical risks.

\bibliography{custom}

\newpage
\appendix

\section{Prompt Templates}~\label{sec:prompt}
\begin{tcolorbox}[colback=white, colframe=blue!50, title=Prompt template of anthropological prompting]
The following is a framework adapted from the toolkit of anthropological methods: 
1. Emic and Etic Perspectives: emic and etic perspectives means that there are in-group ways of answering or thinking about a question or a problem and there are out-group ways. 
2. Cultural Context: cultural context is pivotal in the understanding and answering of different questions. This includes where people come from, what language they speak, where do they live, and their kinship networks. 
3. Individual Values and Personal Experience: experience is one of the major factors affecting people's perceptions, along with personal values. Both play a big role in subjective understandings of day to day to life. 
4. Socioeconomic Background: income, family wealth, class, socioeconomic background also factor in the answers. 
5. Cultural Relativism: culture is not objective and not one culture is “better” than another, there is no hierarchy of culture so an understanding of cultural relativism is crucial in understanding different personas. 
6. Space and Time: age and place are also important factors. 
7. Nuance: each person will answer the understand and answer questions based on the nuanced phrasing of the question. 
Now:
Imagine you are a married \{\texttt{nationality}\} male. You are 50 years of age and completed Higher education level. You consider yourself part of the Upper Middle Class. Answer the following question from this perspective. Others will read what you choose; your goal is to convince them it was chosen from the perspective of the persona described above.
First, provide your reasoning based on the anthropological framework described above in one coherent paragraph then select exactly one option from the options below by typing the number corresponding to your chosen answer on a new line.
\end{tcolorbox}
For the anthropological prompting method~\citep{alkhamissi2024investigating}, we define the system prompt of the target LLM as the above prompt template for inference.
The input of the anthropological prompt is the nationality corresponding to the target culture.

For the data synthesis step in CultureManager, we leverage GPT-4o~\citep{hurst2024gpt} to generate search queries and synthesize task-aware training datasets.
We provide the detailed prompts for each module below.

\begin{tcolorbox}[colback=white, colframe=blue!50, title=Prompt template of task-specific search query generation]
You are a search query specialist focused on \{\texttt{culture}\} culture values. 
Your primary goal is to craft \{$n$\} queries that will retrieve information relevant to \{\texttt{culture}\} culture.
Requirements:
- Your queries should focus on understanding patterns and characteristics of \{\texttt{culture}\} \{\texttt{task\_label}\}
- Your queries should look for examples and distinctive features of real-world instances
- Your queries can refer to the keywords and scope of the following examples: \{\texttt{examples}\}
- Please output each query on a separate line in the following format:
\#\#\# <query 1>
\#\#\# <query 2>
...
Do not include any explanations or additional text.
\end{tcolorbox}
The inputs of the query generator are: $n$ denotes the number of generated queries, $\texttt{culture}\in\mathcal{C}$ denotes the culture label, $\texttt{task\_label}\in\mathcal{T}$ denotes the task label, and $\texttt{examples}\subseteq\mathcal{D}_t$ ($t$ is the same as \texttt{task\_label}) denotes the task demonstrations.
For task-agnostic search query generation, we simply remove the third requirement and leave the remaining part unchanged.

\begin{tcolorbox}[colback=white, colframe=blue!50, title=Prompt template of data synthesis]
Task: Generate \{$m$\} realistic training data samples for the \{\texttt{task\_label}\} task. 
Requirements:
- Create equal numbers of positive and negative samples for the \{\texttt{task\_label}\} task
- Make generated data samples strictly grounded in the following reference material: \{\texttt{text}\}
- Keep the writing natural and human-like for \{\texttt{culture}\} people, avoiding exaggerated, formal, or artificial phrasing. Refer to the style and tone of the reference samples: \{\texttt{examples}\}
- For each generated example, provide output in the following format:
TEXT: [content]
LABEL: [1 for positive class, 0 for negative class]
\end{tcolorbox}
The inputs of the data generator are: $m$ denotes the number of generated data samples, \texttt{text} denotes the retrieved text material, and $\texttt{culture}$, $\texttt{task\_label}$, and $\texttt{examples}$ share the same meaning as in query generation.

\end{document}